\documentclass[10pt,twocolumn,letterpaper]{article}

\usepackage{cvpr}
\usepackage{times}
\usepackage{epsfig}
\usepackage{graphicx}
\usepackage{booktabs}
\usepackage{amsmath}
\usepackage{amssymb}
\usepackage{multicol}
\usepackage{multirow}
\usepackage{float}
\usepackage{multirow}
\usepackage{url}

\usepackage{makecell}
\usepackage{subcaption}
\usepackage{cuted}
\usepackage[font={small}]{caption}
\usepackage{pifont}
\usepackage[dvipsnames]{xcolor}
\usepackage{balance}
\usepackage{epigraph}

\usepackage{algorithm}
\usepackage{algorithmicx}
\usepackage{algpseudocode}
\usepackage{nopageno}

\definecolor{citecolor}{RGB}{65,105,225}
\usepackage[pagebackref=true,breaklinks=true,letterpaper=true,colorlinks,citecolor=citecolor,bookmarks=false]{hyperref}

\cvprfinalcopy 

\def\cvprPaperID{****} 

\begin{document}

\title{TransMoMo: Invariance-Driven Unsupervised Video Motion Retargeting}

\author{Zhuoqian Yang$^{1}\thanks{Equal contribution.}$\qquad Wentao Zhu$^{2*}$\qquad Wayne Wu$^{3*}$ \\ \qquad Chen Qian$^{4}$ \qquad Qiang Zhou$^{3}$\qquad Bolei Zhou$^{5}$\qquad Chen Change Loy$^{6}$ \\
$^1$Robotics Institute, Carnegie Mellon University \hspace{10pt} $^2$Peking University \hspace{10pt} $^3$BNRist, Tsinghua University \\ $^4$SenseTime Research \hspace{10pt} $^5$CUHK \hspace{10pt} $^6$Nanyang Technological University\\
{\tt\small zhuoqiay@cs.cmu.edu}\hspace{1cm}
{\tt\small wtzhu@pku.edu.cn}\hspace{1cm} \\
{\tt\small wwy15@mails.tsinghua.edu.cn}\hspace{1cm}
{\tt\small qianchen@sensetime.com}\hspace{1cm} \\
{\tt\small zhouqiang@tsinghua.edu.cn}\hspace{1cm}
{\tt\small bzhou@ie.cuhk.edu.hk}\hspace{1cm}
{\tt\small ccloy@ntu.edu.sg}
\vspace{-0.5in}
}

\maketitle

\begin{strip}
\centering
\includegraphics[width=0.95\textwidth]{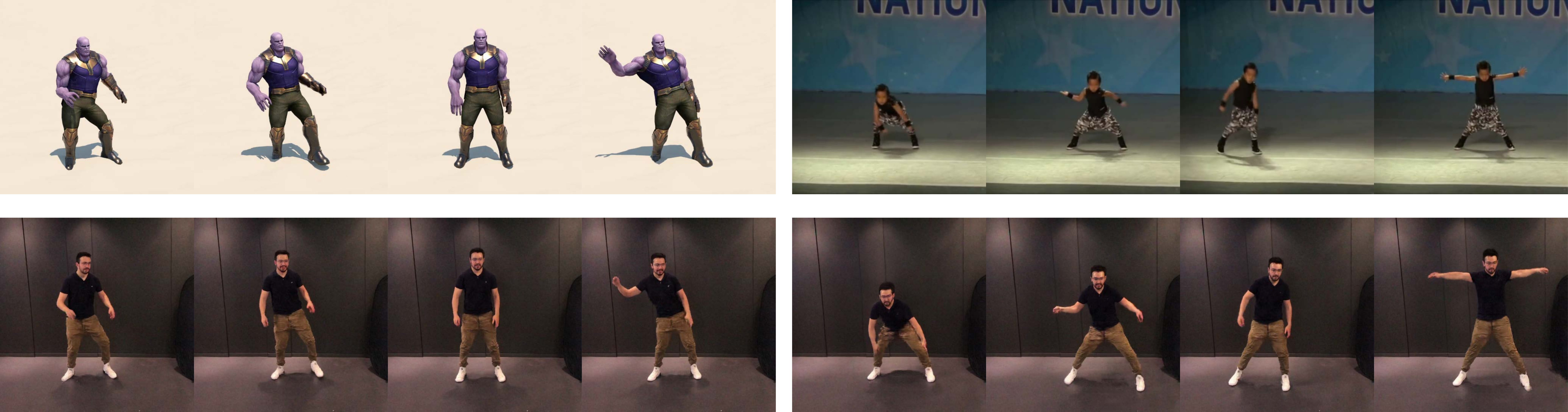}
\captionof{figure}{\textbf{Motion retargeting.} The movements from the source videos (first row) are transferred to a target appearance (second row).}
\vspace{-0.2cm}
\label{fig:fig_intro}
\end{strip}

\begin{abstract}
\vspace{-0.45cm}
We present a lightweight video motion retargeting approach TransMoMo that is capable of transferring motion of a person in a source video realistically to another video of a target person (Fig.~\ref{fig:fig_intro}). Without using any paired data for supervision, the proposed method can be trained in an unsupervised manner by exploiting invariance properties of three orthogonal factors of variation including motion, structure, and view-angle.
Specifically, with loss functions carefully derived based on invariance, we train an auto-encoder to disentangle the latent representations of such factors given the source and target video clips. This allows us to selectively transfer motion extracted from the source video seamlessly to the target video in spite of structural and view-angle disparities between the source and the target.
The relaxed assumption of paired data allows our method to be trained on a vast amount of videos needless of manual annotation of source-target pairing, leading to improved robustness against large structural variations and extreme motion in videos. 
We demonstrate the effectiveness of our method over the state-of-the-art methods such as NKN~\cite{NKN}, EDN~\cite{EDN} and LCM~\cite{LCM}. Code, model and data are publicly available on our project page.\footnote{\color{red}{\url{https://yzhq97.github.io/transmomo}}}
\vspace{-0.5cm}
\end{abstract}

\section{Introduction}


\epigraph{Let's sway you could look into my eyes. Let's sway under the moonlight, this serious moonlight.}{\textit{David Bowie, Let's Dance}}
\vspace{0.1cm}

Can an amateur dancer learn instantly how to dance like a professional in different styles, \eg, Tango, Locking, Salsa, and Kompa?
While it is almost impossible in reality, one can now achieve this virtually via motion retargeting - transferring the motion of a source video featuring a professional dancer to a target video of him/herself.

Motion retargeting is an emerging topic in both computer vision and graphics due to its wide applicability to content creation. Most existing methods~\cite{NKN, Liu2018Neural, lwb2019} achieve motion retargeting through high-quality 3D pose estimation or reconstruction~\cite{kwanyee2019weakly3dpose}. These methods either require complex and expensive optimization or they are error-prone given the unconstrained videos that contain complex motion. 
Recently, several efforts are also made to retarget motion in 2D space~\cite{LCM, EDN, DBLP:conf/cvpr/JooKK18}. Image-based methods~\cite{DBLP:conf/cvpr/EsserSO18, DBLP:conf/cvpr/BalakrishnanZDD18} obtain compelling results on conditional person generation. However, these methods always neglect the temporal coherence in video and thus suffer from twinkling results. Video-based methods~\cite{wang2018vid2vid,EDN,LCM} show state-of-the-art results. However, insufficient consideration of variances between two individuals~\cite{wang2018vid2vid,EDN} or the limitation of training on synthesized data~\cite{LCM} makes their result deteriorate dramatically while encountering large structure variations or extreme motion in web videos.

\begin{figure*}[t]
\centering
\includegraphics[width=0.8\linewidth]{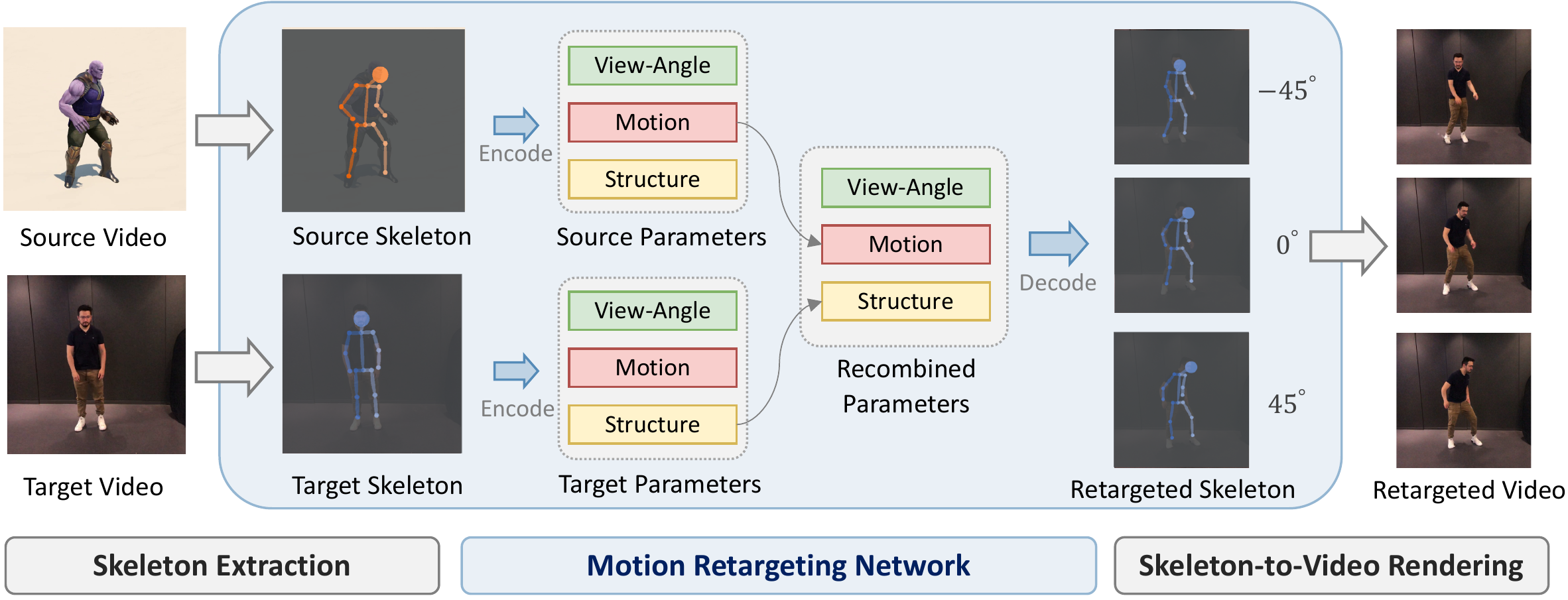}
\vspace{-0.15cm}
\caption{\textbf{Motion retargeting pipeline} Our method achieves motion retargeting in three stages. 1.\textit{Skeleton Extraction}: 2D body joints are extracted from source and target videos using an off-the-shelf model. 2.\textit{Motion Retargeting Network}: our model decomposes the joint sequences and recombines the elements to generate a new joint sequence, which can be viewed at any desired view-angle. 3.\textit{Skeleton-to-Video Rendering}: Retargeted video is rendered using the output joint sequence, with an available image-to-image translation method. }
\label{fig:architecture_inference}
\vspace{-0.35cm}
\end{figure*}

In this study, we aim to address video motion retargeting via an end-to-end learnable framework in 2D space, bypassing the need for explicit estimation of 3D human pose.
Despite recent progress in generative frameworks and motion synthesis, learning for motion retargeting in 2D space remains challenging due to the following issues:
1) Consider the large structural and view-angle variances between the source and target videos, it is difficult to learn a direct person-to-person mapping at the pixel level. Conventional image-to-image translation methods tend to generate unnatural motion in extreme conditions or fail on unseen examples;
2) No corresponding image pairs of two different subjects performing the same motion are available to supervise the learning of such a transfer;
3) Human motion is highly articulated and complex, thus it is challenging to perform motion modeling and transfer.

To address the first challenge, instead of performing direct video-to-video translation at the pixel level, we decompose the translation process into three steps as shown in Fig.~\ref{fig:architecture_inference}, \ie, skeleton extraction, motion retargeting on skeleton and skeleton-to-video rendering. The decomposition allows us to focus on the core problem of motion re-targeting using skeleton sequences as the input and output spaces.
To cope with the second and third challenges, we exploit the invariance property of three factors: motion, structure, and view-angle. These factors of variation are enforced to be independent of each other, held constant when other factors vary. 
In particular, 1) \textit{motion} should be invariant despite structural and view-angle perturbations, 2) \textit{structure} of one skeleton sequence should be consistent across time and invariant despite view-angle perturbations, and 3) \textit{view-angle} of one skeleton sequence should be consistent across time and invariant despite structural perturbations.
The invariance properties allow us to derive a set of purely unsupervised loss functions to train an auto-encoder for disentangling a sequence of skeletons into orthogonal latent representations of motion, structure, and view-angle. Given the disentangled representation, one can easily mix the latent codes of motion and structure from different skeleton sequences for motion retargeting. Taking different view-angle as a condition to the decoder, one can generate retargeted motion in novel viewpoints. Since motion retargeting is performed on the 2D skeleton space, it can be seen as a lightweight and plug-and-play module, which is complementary to existing skeleton extraction~\cite{cao2018openpose, alp2018densepose, peng2018jointly, DBLP:conf/iccv/YangLOLW17} and skeleton-to-video rendering methods~\cite{EDN, wang2018vid2vid, wang2019fewshotvid2vid}.


There are several existing studies designed for \textit{general} representation disentanglement in video~\cite{DBLP:conf/nips/HsiehLHLN18, DBLP:conf/cvpr/Tulyakov0YK18, DBLP:conf/nips/DentonB17}. While these methods have shown impressive results on constrained scenarios. It is difficult for them to model articulated human motion due to the highly non-linear and complex kinematic structures. Instead, our method is designed \textit{specifically} for representation disentanglement in human videos.

We summarize our contributions as follows: 
1) We propose a novel Motion Retargeting Network in 2D skeleton space, which can be trained end-to-end with unlabeled web data.
2) We introduce novel loss functions based on invariance to endow the proposed network with the ability to disentangle representation in a purely unsupervised manner.
3) Extensive experiments demonstrate the effectiveness of our method over other state-of-the-art approaches~\cite{EDN,LCM,NKN}, especially under in-the-wild scenarios where motion are complex.

\section{Related Work}

\noindent \textbf{Video Motion Retargeting.} Hodgins and Pollard~\cite{DBLP:conf/siggraph/HodginsP97} proposed a control system parameter scaling algorithm to adapt simulated motion to new characters. Lee and Shin~\cite{DBLP:conf/siggraph/LeeS99} decomposed the problem into inter-frame constraints and intra-frame relationships and modeled them by Inverse Kinematics problem and B-spline curve separately. Choi and Ko \cite{DBLP:journals/jvca/ChoiK00} proposed a real-time method based on inverse rate control that computes the changes in joint angles. Tak and Ko \cite{DBLP:journals/tog/TakK05} proposed a per-frame filter framework to generate physically plausible motion sequences. Recently, Villegas \etal \cite{NKN} designed a recurrent neural network architecture with a Forward Kinematics layer to capture high-level properties of motion. However, the target to be animated of the aforementioned approaches is typically an articulated virtual character and their results critically depend on the accuracy of 3D pose estimation. More recently, Aberman \etal \cite{LCM} propose to retarget motion in 2D space. However, since their training relies on synthetic paired data, the performance is likely to degrade under the unconstrained scenarios. Instead, our method can be trained on pure unlabeled web data, which makes the method robust to the challenging in-the-wild motion transfer task.

There exist a few attempts to address the video motion retargeting problem. Liu \etal \cite{Liu2018Neural} designed a novel GAN~\cite{goodfellowgan} architecture with an attentive discriminator network and better conditioning inputs. However, this method relies on 3D reconstruction of the target person. Aberman \etal \cite{DBLP:journals/cgf/AbermanSLLCC19} proposed to tackle video-driven performance cloning in a two-branch framework. Chan \etal \cite{EDN} proposed a simple but effective method to obtain temporal coherent video results. Wang \etal \cite{wang2018vid2vid} achieves results of similar quality to Chan \etal with more complex shape representation and temporal modelling. However, The performance of all these methods degrades dramatically when large variations happened between two individuals with no consideration~\cite{DBLP:journals/cgf/AbermanSLLCC19, wang2019fewshotvid2vid, wang2018vid2vid} or a simple rescaling~\cite{EDN} to address body variations.



\noindent \textbf{Unsupervised Representation Disentanglement.} 
There is a vast literature~\cite{DRIT, liu2019few, huang2018munit,shengju2019makeaface,wayne2019transgaga,wayne2019disentangling} on disentangling factors of variation. Bilinear models~\cite{DBLP:journals/neco/TenenbaumF00} were an early approach to separate content and style for images of faces and text in various fonts. Recently, InfoGAN~\cite{chen2016infogan} learned a generative model with disentangled factors based on Generative Adversarial Networks (GAN). $\beta$-VAE~\cite{higgins2017beta} and DIP-VAE~\cite{DBLP:conf/iclr/0001SB18}, build on variational Auto-Encoders (VAEs) to disentangle interpretable factors in an unsupervised way.

Other approaches explore general methods for learning disentangled representations from video. Whitney \etal~\cite{DBLP:journals/corr/WhitneyCKT16} used a gating principle to encourage each dimension of the latent representation to capture a distinct mode of variation. Villegas \etal~\cite{DBLP:conf/iclr/VillegasYHLL17} used an unsupervised approach to factoring video into content and motion. Denton~\etal~\cite{DBLP:conf/nips/DentonB17} proposed to leverage the temporal coherence of video and a novel adversarial loss to learn a disentangled representation. MoCoGAN~\cite{DBLP:conf/cvpr/Tulyakov0YK18} employs unsupervised adversarial training to learn the separation of motion and content. Hsieh~\etal~\cite{DBLP:conf/nips/HsiehLHLN18} proposed an auto-encoder framework, which combines structured probabilistic models and deep networks for disentanglement. However, the performance of these methods are not satisfactory on human videos, since they are not designed specifically for disentanglement of highly articulated and complex objects.


\noindent \textbf{Person Generation.} Various machine learning algorithms have been used to generate realistic person images. The generation process could be conditionally guided by skeleton keypoints~\cite{DBLP:conf/cvpr/BalakrishnanZDD18, DBLP:conf/nips/MaJSSTG17} and style codes~\cite{DBLP:conf/cvpr/MaSGGSF18, DBLP:conf/cvpr/EsserSO18, DBLP:conf/wacv/BemGBAST19}. Our method is complementary to the image-based person generation approaches and can further boost the temporal coherence of them since it performs motion retargeting on the 2D skeletons space only.
















\section{Methodology}

As illustrated in Fig.~\ref{fig:architecture_inference}, we decompose the translation process into three steps, \ie, skeleton extraction, motion retargeting and skeleton-to-video rendering. In our framework, motion retargeting is the most important component in which we introduce our core contribution (\ie, invariance-driven disentanglement).
Skeleton extraction and skeleton-to-video rendering are replaceable and can thus benefit from recent advances in 2D keypoints estimation~\cite{alp2018densepose, cao2018openpose, DBLP:conf/iccv/YangLOLW17} and image-to-image translation~\cite{pix2pix2017, wang2018vid2vid, wang2019fewshotvid2vid}.

The Motion Retargeting Network decomposes 2D joint input sequences as a \emph{motion} code that represents the movements of the actor, a \emph{structure} code that represents the body shape of the actor and a \emph{view-angle} code that represents the camera angle. The decoder takes any combination of the latent codes and produces a reconstructed 3D joint sequence, which automatically isolates view from motion and structure. 


For transferring motion from a source video to a target video, we first use an off-the-shelf 2D keypoints detector to extract joint sequences from videos. By combining the motion code encoded from the source sequence and the structure code encoded from the target sequence, our model then yields a transferred 3D joint sequence. The transferred sequence is then projected back to 2D with any desired view-angle. Finally, we convert the 2D joint sequence frame-by-frame to a pixel-level representation, \ie, label maps. These label maps are fed into a pre-trained image-to-image generator to render the transferred video.

\subsection{Motion Retargeting Network}

Here, we detail the encoders and decoders for an input sequence $\mathbf{x} \in \mathbb{R}^{T \times 2N}$ where $T$ is the length of the sequence and $N$ is the number of body joints.

The motion encoder uses several layers of one dimensional temporal convolution to extract motion information: $E_m(\mathbf{x})=\mathbf{m} \in \mathbb{R}^{M \times C_m}$, where $M$ is the sequence length after encoding and $C_m$ is the number of channels. Note that the motion code $\mathbf{m}$ is variable in length so as to preserve temporal information. 

The structure encoder has a similar network structure $E_s(\mathbf{x})=\mathbf{s}\in \mathbb{R}^{M \times C_s}$, with the difference that the final structure code is obtained after a temporal max pooling: $\bar{E_s}(\mathbf{x})=\bar{\mathbf{s}}=\text{maxpool}(\mathbf{s})$, therefore $\bar{\mathbf{s}} \in \mathbb{R}^{C_s}$.
Effectively, the process of obtaining the structure code can be interpreted as performing multiple body shape estimations in sliding windows: $E_s(\mathbf{x})=[\mathbf{s}_1, \mathbf{s}_2, ..., \mathbf{s}_M]$, and then aggregating the estimations.
Assuming the viewpoint is also stationary (i.e. all the temporal variances are caused by the movements of the actor), the view code $\bar{E_v}(\mathbf{x}) = \bar{\mathbf{v}} \in \mathbb{R}^{C_v}$ is obtained the same way we obtained the structure code.

The decoder takes the motion, body and view codes as input and reconstructs a 3D joint sequence $G(\mathbf{m}, \bar{\mathbf{s}}, \bar{\mathbf{v}})=\hat{\mathbf{X}} \in \mathbb{R}^{T \times 3N}$ through convolution layers, in symmetry with the encoders. Our discriminator $D$ is a temporal convolutional network that is similar to our motion encoder: $D(\mathbf{x}) \in \mathbb{R}^{M}$.

\subsection{Invariance-Driven Disentanglement}

The disentanglement of motion, structure and view is achieved leveraging the invariance of each of these factors to the changes in the other twos. We design loss terms to restrict changes when perturbations are added, while the entire network tries to reconstruct joint sequences from decomposed features. Structural perturbation is added through \emph{limb scaling}, i.e. manually shortening or extending the length of the limbs. View perturbation is introduced through rotating the reconstructed 3D sequence and projecting it back to 2D. Motion perturbation needs not be explicitly added since motion itself is varying through time. We first describe the ways perturbations are added and then detail the definitions of the loss terms derived from three invariances, \ie, motion, structure and view-angle invariance.

\begin{figure}[t]
\vspace{-0.25cm}
\centering
\includegraphics[width=0.95\linewidth]{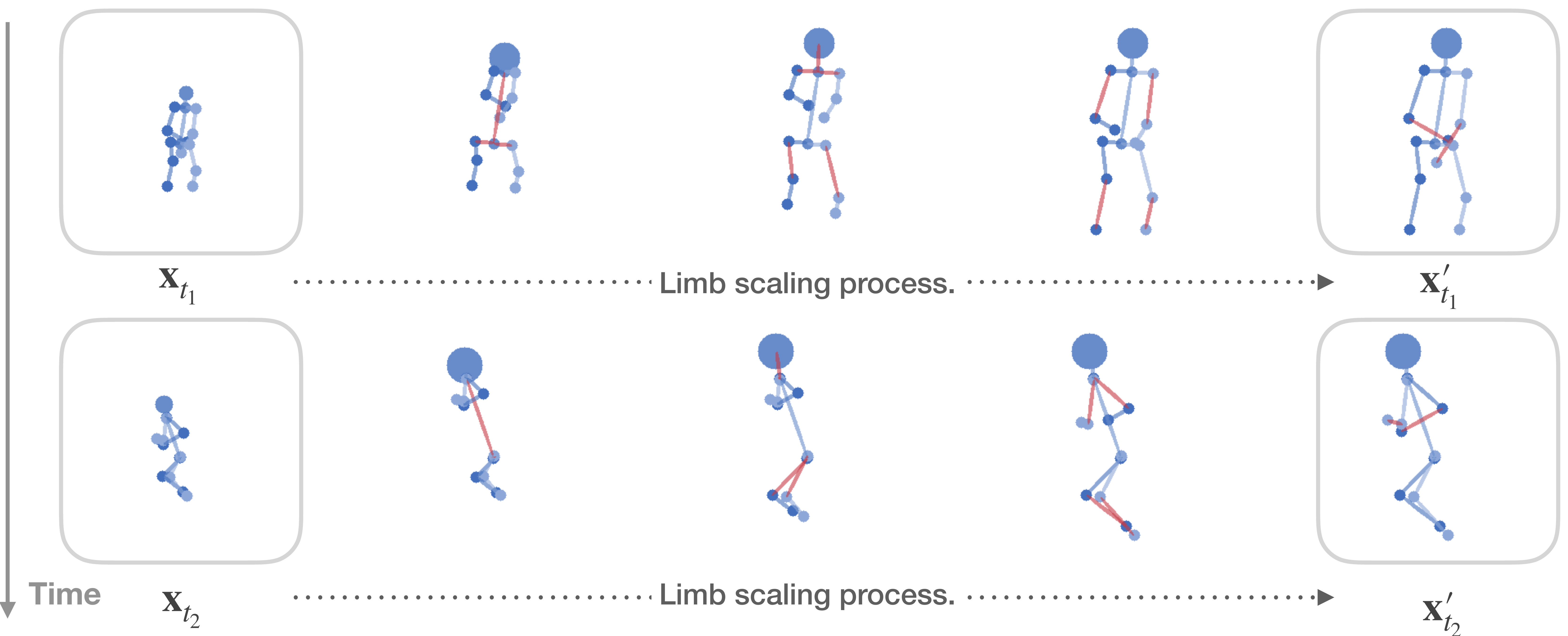}
\vspace{-0.25cm}
\caption{\textbf{Limb-scaling process.} We show a step-by-step limb-scaling process on a joint sequence $\mathbf{x}$ starting from the root joint (pelvis). At each step, the scaled limbs are highlighted with red. This example scales all limbs with the same factor $\gamma_i=2$, but the scaling factors are randomly generated at training time.}
\label{fig:fig_limb_scale}
\vspace{-0.35cm}
\end{figure}

\noindent \textbf{Limb Scaling as Structural Perturbation.} For an input 2D sequence $\mathbf{x} \in \mathbb{R}^{T \times 2N}$, we create a structurally-perturbed sequence by elongating or shortening the limbs of the performer, as illustrated in Figure~\ref{fig:fig_limb_scale}. It is done in such a way that the created sequence is effectively the same motion performed by a different actor. The length of a limb is extended/shortened by the same ratio across all frames, so limb-scaling does not introduce ambiguity between motion and body structure. Specifically, the limb-scaled sequence $\mathbf{x}^\prime$ is created by applying the limb-scale function frame-by-frame: $\mathbf{x}_t^\prime = \delta(\mathbf{x}_t; \boldsymbol{\gamma}, \gamma_g)$, where $\mathbf{x}_t$ is the $t^\text{th}$ frame in the input sequence, $\delta$ is the limb scaling function, $\boldsymbol{\gamma}=[\gamma_1, \gamma_2, ...]$ are the local scaling factors and $\gamma_g$ is the global scaling factor. Modeling the human skeleton as a tree and joints as its nodes, we define the pelvis joint as the root. For each frame in the sequence, starting from the root, we recursively move the joints and all their dependent joints (child nodes) on the direction of the limb by distance $(\gamma_i-1)L_i^{(t)}$, where $L_i^{(t)}$ is the original length of the limb in the $t^\text{th}$ frame.
After all local scaling factors have been applied, the global scaling factor $\gamma_g$ is directly multiplied with all the joint coordinates.

\noindent \textbf{3D Rotation as View Perturbation.}
Let $\phi$ be a rotate-and-project function, \ie, for a 3D coordinate $\mathbf{p}=[x~y~z]^T$:
\[
    \phi(\mathbf{p}, \theta; \mathbf{n}) =
    \begin{bmatrix}
    R_{11}(\mathbf{n}, \theta) & R_{12}(\mathbf{n}, \theta) & R_{13}(\mathbf{n}, \theta) \\
    R_{21}(\mathbf{n}, \theta) & R_{22}(\mathbf{n}, \theta) & R_{23}(\mathbf{n}, \theta) \\
    \end{bmatrix}
    \begin{bmatrix}
    x \\ y \\ z
    \end{bmatrix}
\]
$R(\mathbf{n}, \theta) \in \mathbb{SO}^3$ is a rotation matrix obtained using Rodrigues' rotation formula and $\mathbf{n}$ is a unit vector representing the axis around which we rotate. In practice, $\mathbf{n}$ is an estimated vertical direction of the body. It is computed using four points: left shoulder, right shoulder, left pelvis and right pelvis.
Note that $\phi(\mathbf{p}, \theta)$ is differentiable with respect to $\mathbf{p}$.

\begin{figure}[t]
\vspace{-0.15cm}
\centering
\includegraphics[width=1.0\linewidth]{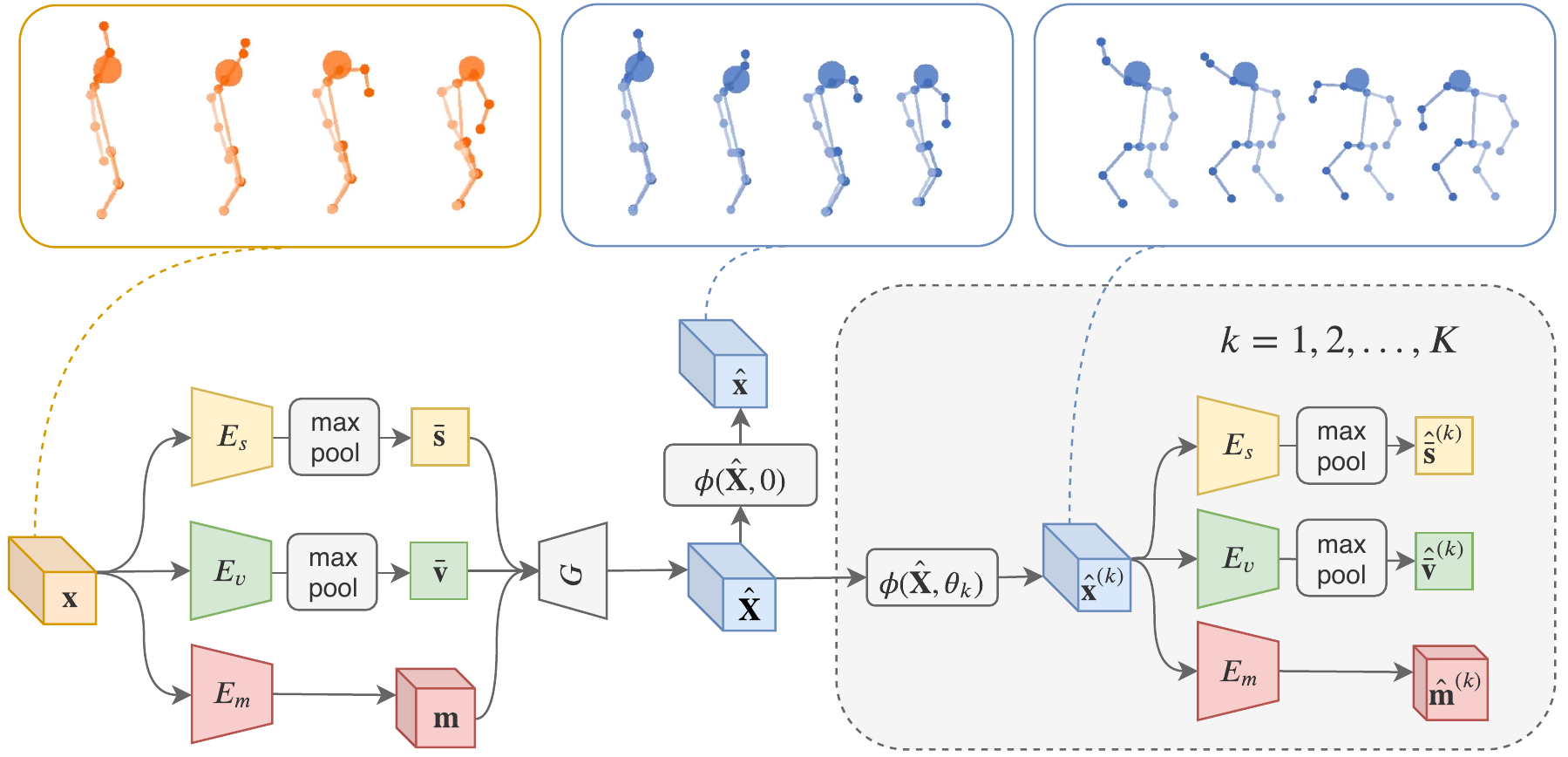}
\vspace{-0.45cm}
\caption{\textbf{Rotation as view perturbation.} This figure illustrates the process of taking an input 2D sequence $\mathbf{x}$, reconstructing a 3D sequence $\hat{\mathbf{X}}$ using our motion retargeting network and projecting it back to 2D, with rotation as view-angle perturbation.}
\label{fig:fig_rotate}
\vspace{-0.45cm}
\end{figure}

As shown in Fig.~\ref{fig:fig_rotate}, we create several rotated sequences from the reconstructed 3D sequence $\hat{\mathbf{X}}$:
\[\hat{\mathbf{x}}^{(k)} = \phi(\hat{\mathbf{X}}, \frac{k}{K+1}\pi),~k=1,2,...,K \]
and $K$ is number of projections. Loss terms enforcing disentanglement will be described later in this chapter.


\subsubsection{Invariance of Motion} 

Motion should be invariant despite structural and view-angle perturbations. To this end, we designed the following loss terms.

\begin{figure}[t]
\centering
\includegraphics[width=0.9\linewidth]{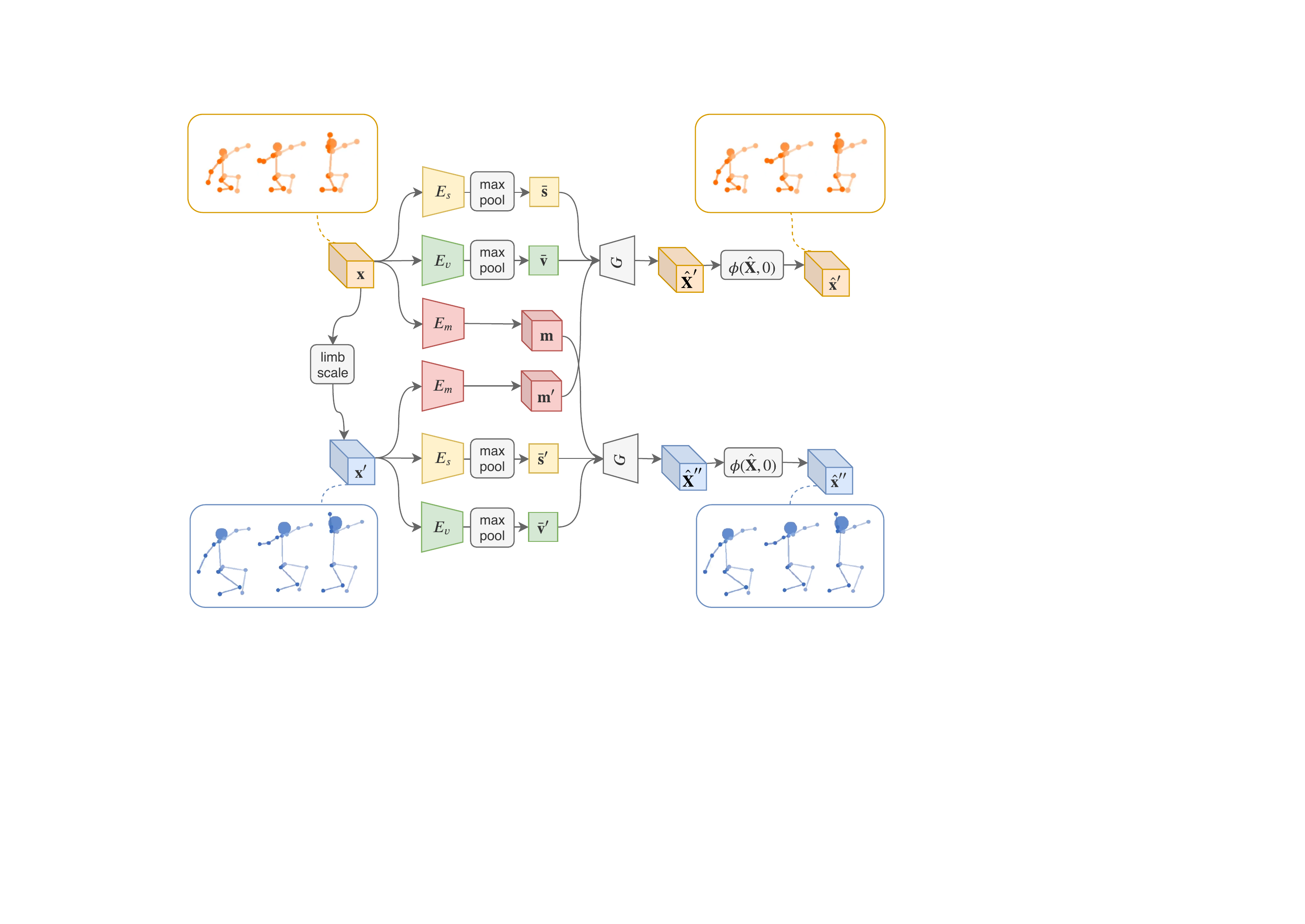}
\vspace{-0.25cm}
\caption{\textbf{Cross reconstruction process.} This figure illustrates the process of cross-reconstruction using a 2D input sequence $\mathbf{x}$ and its limb-scaled variant $\mathbf{x}^\prime$. }
\label{fig:fig_cross}
\vspace{-0.45cm}
\end{figure}

\noindent \textbf{Cross Reconstruction Loss.} Recall that we use limb scaling to obtain data of the same movements performed by ``different'' actors $\mathbf{x}$ and $\mathbf{x}^\prime$. We cross reconstruct the two sequences, as shown in Fig.\ref{fig:fig_cross}. The cross reconstruction involves encoding, swapping and decoding, namely: 
\begin{align*}
    \hat{\mathbf{x}}^\prime &= \phi \left[G(E_m(\mathbf{x}^\prime), \bar{E}_s(\mathbf{x}), \bar{E}_v(\mathbf{x})), 0 \right] \\
    \hat{\mathbf{x}}^{\prime\prime} &=\phi \left[G(E_m(\mathbf{x}), \bar{E}_s(\mathbf{x}^\prime), \bar{E}_v(\mathbf{x}^\prime)), 0\right],
\end{align*}
where $\mathbf{x}^\prime$ is the limb-scaled version of $\mathbf{x}$.
Since $\mathbf{x}$ and $\mathbf{x}^\prime$ have the same motion, we expect $\hat{\mathbf{x}}^\prime$ to be the same as $\mathbf{x}$; $\hat{\mathbf{x}}^{\prime\prime}$ to be the same as $\mathbf{x}^\prime$. Therefore, the cross reconstruction loss is defined as
\begin{equation}
    \mathcal{L}_\text{crs} =  \frac{1}{2NT} \left( \frac{1}{2} \left| \mathbf{x} - \hat{\mathbf{x}}^\prime \right| + \frac{1}{2} \left| \mathbf{x}^\prime - \hat{\mathbf{x}}^{\prime\prime} \right| \right).
\end{equation}
\noindent \textbf{Structural Invariance Loss.} This signal is to ensure that the motion codes are invariant to structural changes. $\mathbf{x}$ and $\mathbf{x}^\prime$ have the same motion but different body structures, therefore we expect the motion encoder to have the same output:
\begin{equation}
    \mathcal{L}_\text{inv\_m}^{(s)} = \frac{1}{MC_m} \left| E_m(\mathbf{x}) - E_m(\mathbf{x}^{\prime}) \right|. 
\end{equation}\\
\noindent \textbf{Rotation Invariance Loss.} Similarly, to ensure that the motion code is invariant to rotation, we add:
\begin{equation}
    \mathcal{L}_\text{inv\_m}^{(v)} = \frac{1}{KMC_m} \sum_{k=1}^K \left| E_m(\mathbf{x}) - E_m(\hat{\mathbf{x}}^{(k)}) \right|,
\end{equation}
where $\hat{\mathbf{x}}^{(k)}$ is the $k^\text{th}$ rotated variant.

\subsubsection{Invariance of Structure}

Body structure should be consistent across time and invariant view-angle perturbations.\\
\noindent\textbf{Triplet Loss.} The triplet loss is added to exploit the time-invariant property of the body structure and thereby better enforce disentanglement. Recall that the body encoder produces multiple body structure estimations $E_s(\mathbf{x})=[\mathbf{s}_1, \mathbf{s}_2, ..., \mathbf{s}_M],~E_s(\mathbf{x}^\prime)=[\mathbf{s}^\prime_1, \mathbf{s}^\prime_2, ...\mathbf{s}^\prime_M]$ before averaging them. The triplet loss is designed to map estimations from the same sequence to a small neighborhood while alienating estimations from different sequences.
Let us define an individual triplet loss term:
\begin{equation}
\tau(\mathbf{s}_{t_1}, \mathbf{s}_{t_2}, \mathbf{s}^\prime_{t_2}) = \max \left\{0, s(\mathbf{s}_{t_1}, \mathbf{s}^\prime_{t_2}) - s(\mathbf{s}_{t_1}, \mathbf{s}_{t_2})  + m \right\},
\label{eq:trip_term}
\end{equation}
where $s(\cdot, \cdot)$ denotes the cosine similarity function and $m=0.2$ is our margin. The total triplet loss for the invariance of structure is defined as
\begin{equation}
    \mathcal{L}_\text{trip\_s} = \frac{1}{2M} \sum_{t_1, t_2} \left[
    \tau(\mathbf{s}_{t_1}, \mathbf{s}_{t_2}, \mathbf{s}^\prime_{t_2}) +
    \tau(\mathbf{s}^\prime_{t_1}, \mathbf{s}^\prime_{t_2}, \mathbf{s}_{t_2}) \right],
\end{equation}
where $t_1 \neq t_2$.\\
\noindent\textbf{Rotation Invariance Loss.} This signal is to ensure that the structure codes are invariant to rotation:
\begin{equation}
    \mathcal{L}_\text{inv\_s} = \frac{1}{KC_s} \sum_{i=1}^K \left| \bar{E_s}(\mathbf{x}) - \bar{E}_s(\hat{\mathbf{x}}^{(k)}) \right|,
\end{equation}
where $\hat{\mathbf{x}}^{(k)}$ is the $k^\text{th}$ rotated variant.

\subsubsection{Invariance of View-Angle}

View-angle of one skeleton sequence should be consistent through time invariant despite structural perturbations. \\
\noindent\textbf{Triplet Loss.}
Similarly, triplet loss is designed to map view estimations from the same sequence to a small neighborhood while alienating estimations from rotated sequences. Continuing to use the definition of a triplet term in Eq.\ref{eq:trip_term}:
\begin{equation}
    \mathcal{L}_\text{trip} = \frac{1}{2MK} \sum_{k, t_1, t_2} \left[ 
    \tau(\mathbf{v}_{t_1}, \mathbf{v}_{t_2}, \mathbf{v}^{(k)}_{t_2}) +
    \tau(\mathbf{v}^{(k)}_{t_1}, \mathbf{v}^{(k)}_{t_2}, \mathbf{v}_{t_2})
    \right],
\end{equation}
where $\mathbf{v}^{(k)} = \bar{E}_v(\hat{\mathbf{x}}^{(k)})$, $t_1 \neq t_2$.\\
\noindent \textbf{Structural Invariance Loss} This signal is to ensure that the view code is invariant to structural change:
\begin{equation}
    \mathcal{L}_\text{inv\_v} = \frac{1}{C_v} \left| \bar{E}_v(\mathbf{x}) - \bar{E}_v(\mathbf{x}^{\prime}) \right|, 
\end{equation}
where $\mathbf{x}^{\prime}$ is the limb-scaled version of $\mathbf{x}$.

\subsubsection{Training Regularization}

The loss terms defined above are designed to enforce disentanglement. Besides them, some basic loss terms are needed for this representation learning process.\\
\noindent \textbf{Reconstruction Loss.}
Reconstructing data is the fundamental functionality of auto-encoders. Recall that our decoder outputs reconstructed 3D sequences. Our reconstruction loss minimizes the difference between real data and 3D reconstructions projected back to 2D.
\begin{equation}
    \mathcal{L}_\text{rec} = \frac{1}{2NT} \left| \mathbf{x} - \phi(\hat{\mathbf{X}}, 0) \right|,
\end{equation}
i.e. we expect $\hat{\mathbf{X}}$ to be the same as the input $\mathbf{x}$ when we directly remove the $z$ coordinates from $\hat{\mathbf{X}}$. 

\noindent \textbf{Adversarial Loss.}
The unsupervised recovery of 3D motion from joint sequences is achieved through adversarial training. Reconstructed 3D joint sequences are rotated and projected back to 2D and a discriminator is used to measure the domain discrepancy between the projected 2D sequences and real 2D sequences. The feasibility of recovering static 3D human pose from 2D coordinates with adversarial learning has been verified in several works \cite{ramakrishna2012reconstructing,drover2018can, chen2019unsupervised,pavllo20193d}.
We want the reconstructed 3D sequence $\hat{\mathbf{X}}$ to look right after we rotate it and project it back to 2D, therefore the adversarial loss is defined as.
\begin{equation}
    \mathcal{L}_\text{adv} = \frac{1}{K} \sum_{k=1}^K \mathbb{E}_{\mathbf{x} \sim p_\mathbf{x}} [ \frac{1}{2}\log D(\mathbf{x}) + \frac{1}{2} \log(1-D(\hat{\mathbf{x}}^{(k)}))]
\end{equation}

\subsubsection{Total Loss}
The proposed motion retargeting network can be trained end-to-end with a weighted sum of the loss terms defined above:
\begin{align*}
\mathcal{L}
&= \lambda_\text{rec} \mathcal{L}_\text{rec} + \lambda_\text{crs}\mathcal{L}_\text{crs} + \lambda_\text{adv} \mathcal{L}_\text{adv} + \lambda_\text{trip}(\mathcal{L}_\text{trip\_s} + \mathcal{L}_\text{trip\_v}) \\
&+ \lambda_\text{inv} ( \mathcal{L}_\text{inv\_m}^{(s)} +  \mathcal{L}_\text{inv\_m}^{(v)} + 
\mathcal{L}_\text{inv\_s} + \mathcal{L}_\text{inv\_v} )
\end{align*}

\section{Experiments}


\begin{figure}[h]
\begin{center}
\includegraphics[width=0.95\linewidth]{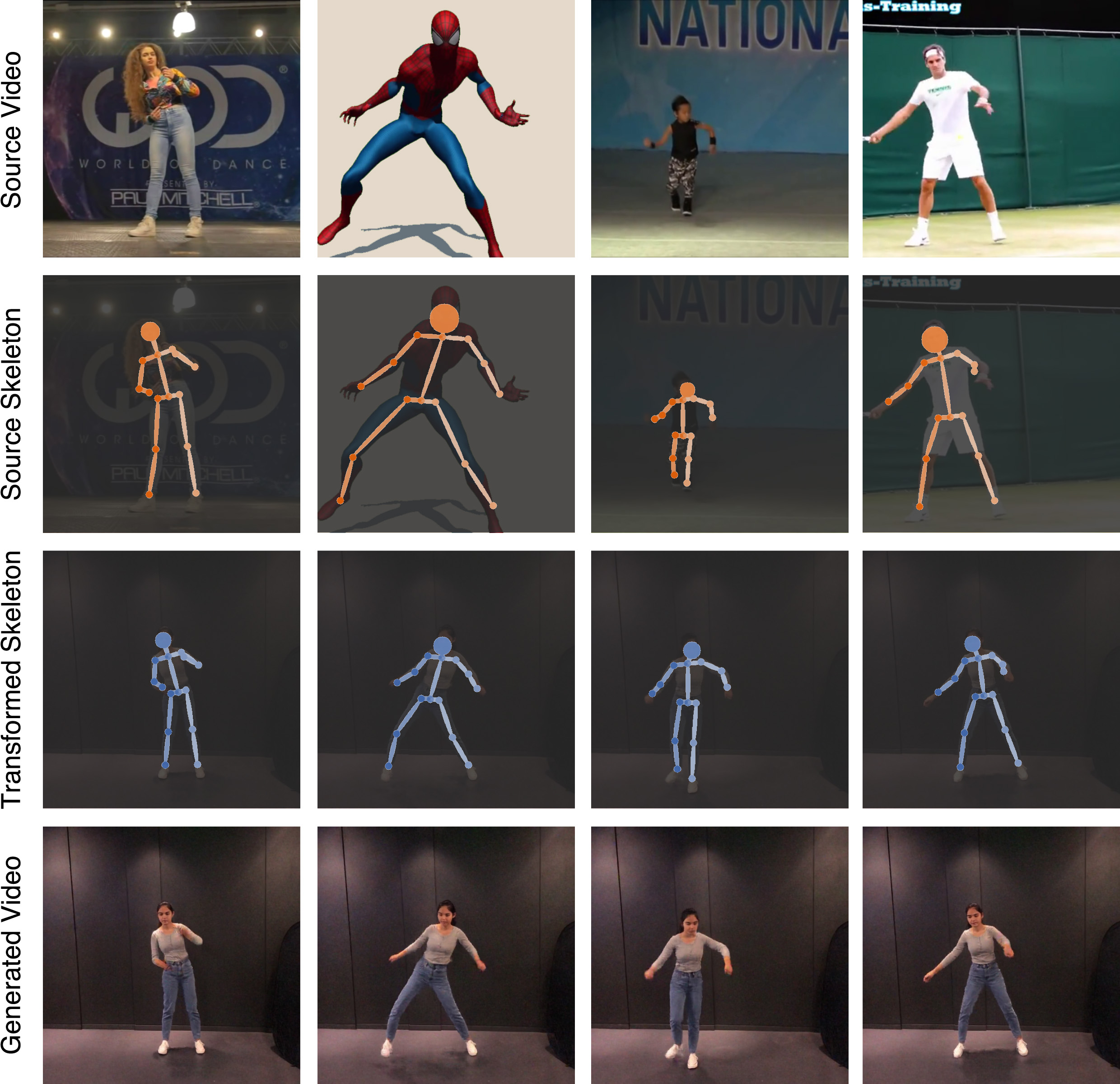}
\end{center}
\vspace{-0.5cm}
\caption{\textbf{Motion retargeting results.} Top to bottom: input source frame, extracted source skeleton, transformed skeleton, generated frame.}
\vspace{-0.45cm}
\label{fig:motion_transfer}
\end{figure}

\begin{figure*}[h]
\vspace{-0.45cm}
\begin{center}
\includegraphics[width=0.9\linewidth]{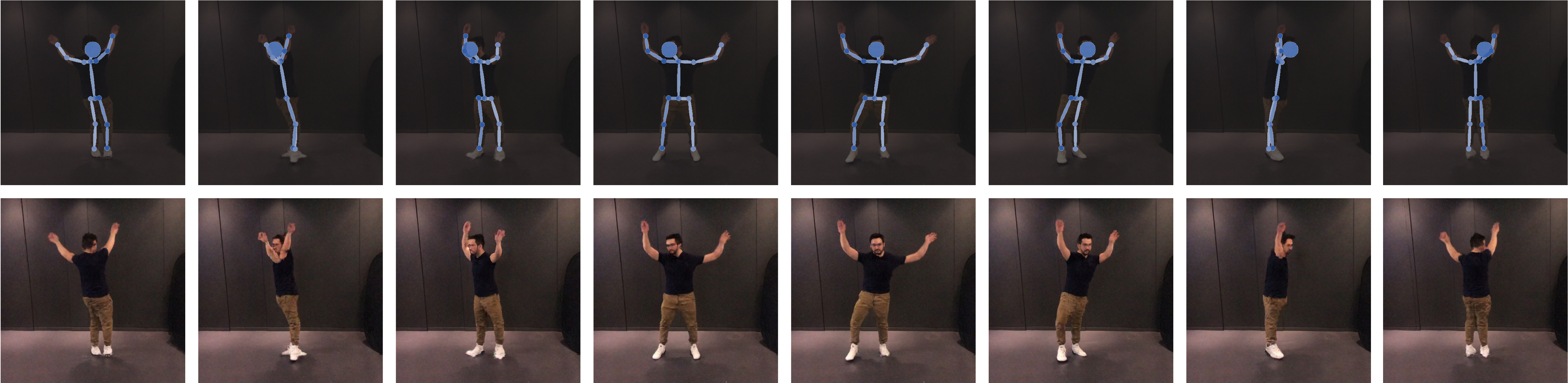}
\end{center}
\vspace{-0.45cm}
\caption{\textbf{Novel view synthesis results.} The first row shows the continuous rotation of generated skeleton, and the second row shows the corresponding rendering results.}
\label{fig:novel_view}
\end{figure*}

\begin{figure}[h]
\vspace{-0.05cm}
\begin{center}
\includegraphics[width=0.95\linewidth]{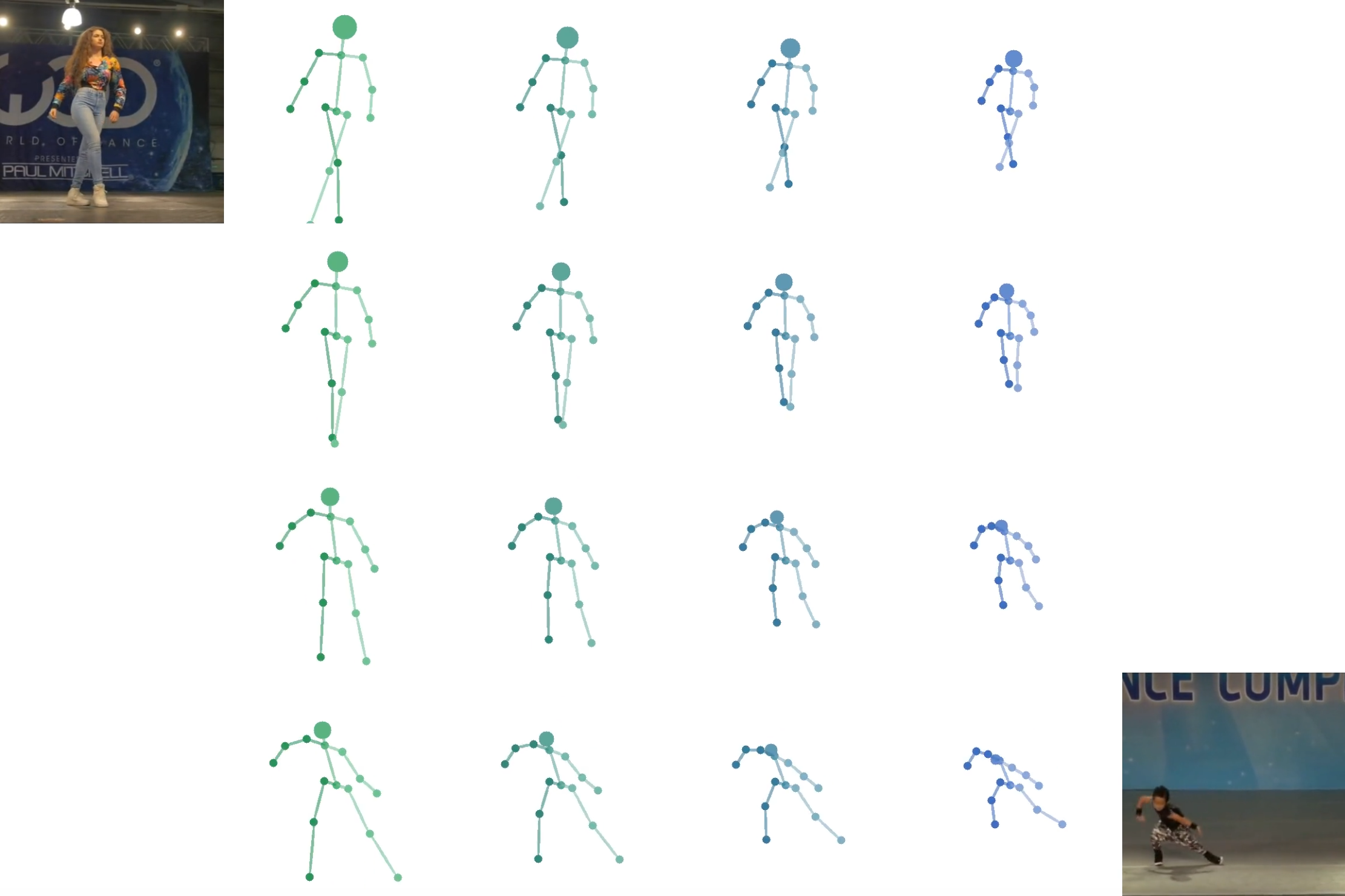}
\end{center}
\caption{\textbf{Latent space interpolation results.} Linear interpolation is tested for body structure (horizontal axis) and motion (vertical axis).}
\label{fig:interpolation}
\vspace{-0.5cm}
\end{figure}

\begin{figure*}[h]
\vspace{-0.25cm}
\begin{center}
\includegraphics[width=0.9\linewidth]{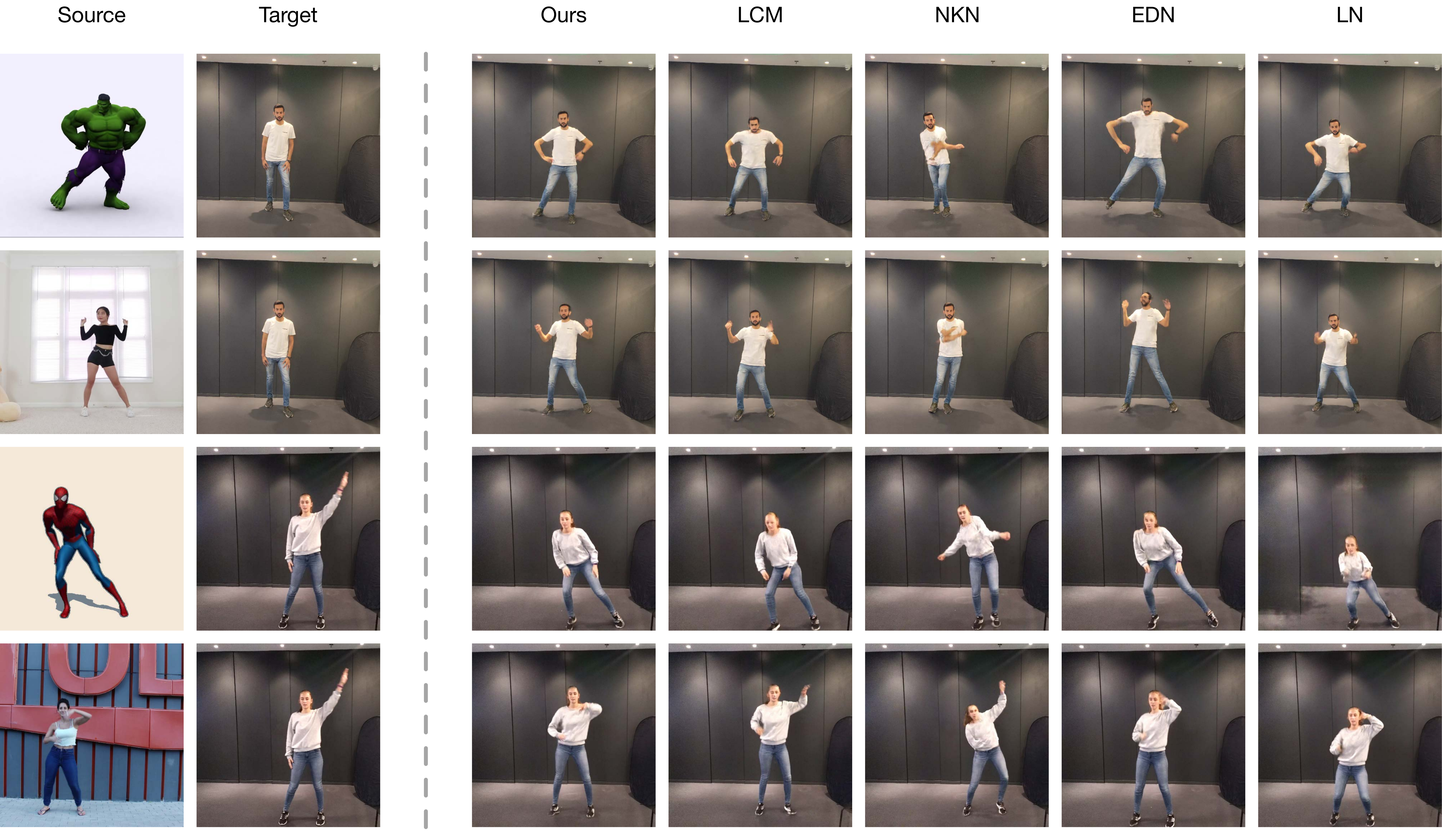}
\end{center}
\vspace{-0.45cm}
\caption{\textbf{Qualitative comparison with state-of-the-art methods.} Each column on the right represents a motion retargeting method.}
\label{fig:comparison_baseline}
\vspace{-0.25cm}
\end{figure*}


\begin{figure}[h]
\vspace{-0.25cm}
\begin{center}
\includegraphics[width=0.85\linewidth]{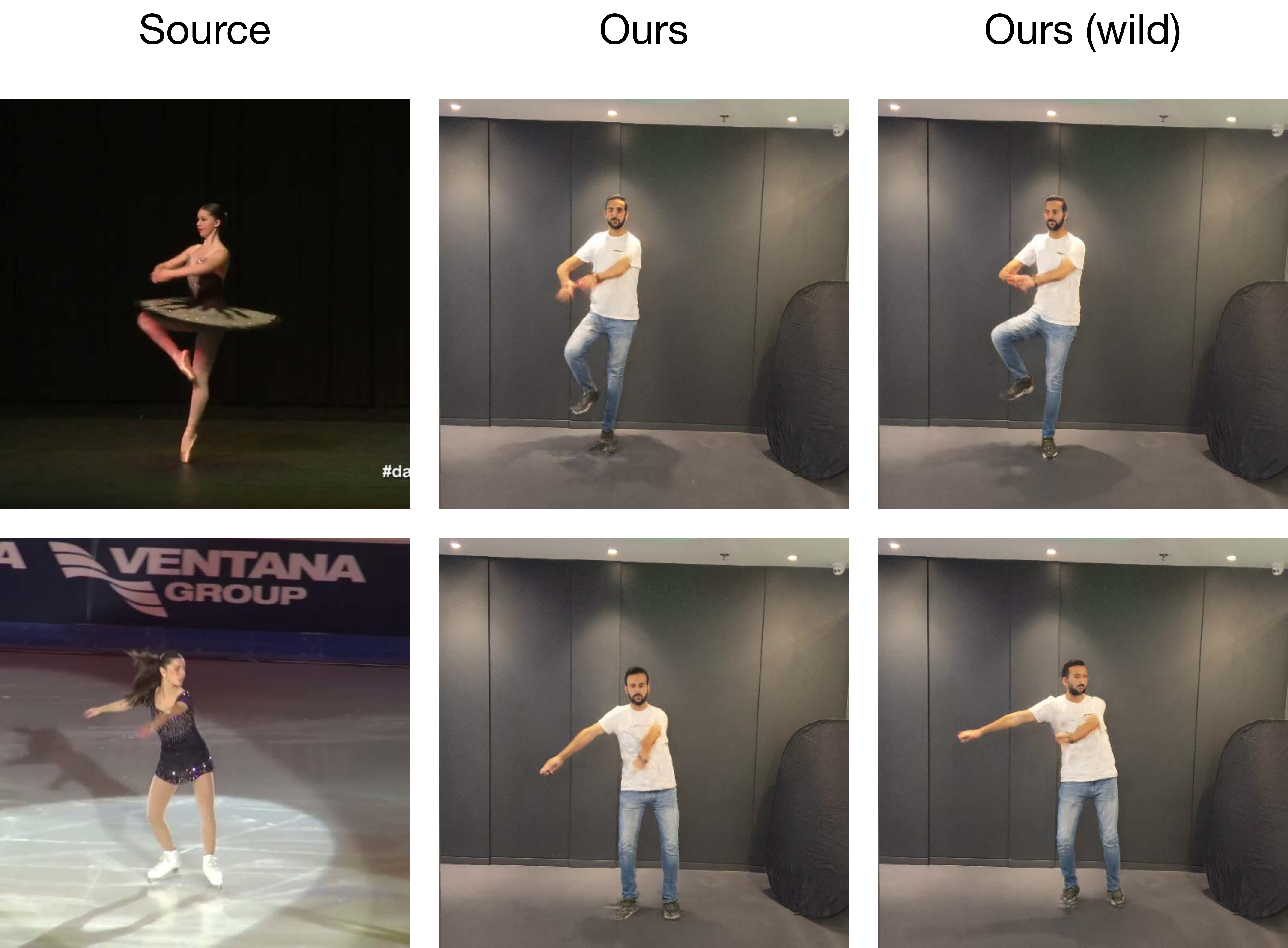}
\end{center}
\vspace{-0.25cm}
\caption{\textbf{Results of our in-the-wild trained model.} Qualitative comparison for models trained with our method on Mixamo and Solo-Dancer separately. The first column gives two challenging motion sources, and the other columns show corresponding results.}
\label{fig:comparison_wild}
\vspace{-0.5cm}
\end{figure}

\subsection{Setup} 

\noindent\textbf{Implementation details.} We perform the proposed training pipeline on the synthetic Mixamo dataset \cite{mixamo} for quantitative error measurement and fair comparison. For in-the-wild training, we collected a motion dataset named Solo-Dancer from online videos. For skeleton-to-video rendering, we recorded $5$ target videos and use the synthesis pipeline proposed in \cite{EDN}. The trained generator is shared by all the motion retargeting methods.

\noindent\textbf{Evaluation metrics.} We evaluate the quality of motion retargeting for both skeleton and video, as retargeting results on skeleton would largely influence the quality of generated videos. For skeleton keypoints, we perform evaluations on a held-out test set from Mixamo (with ground truth available) using mean square error (MSE) as the metric. For generated videos, we evaluate the quality of frames with FID score~\cite{heusel2017gans} and through  a user study.

\subsection{Representation Disentanglement}
We train the model on unconstrained videos in-the-wild, and the model automatically learns the disentangled representations of motion, body, and view-angle, which enables a wide range of applications. We test motion retargeting, novel-view synthesis and latent space interpolation to demonstrate the effectiveness of the proposed pipeline.    
\noindent\textbf{Motion retargeting.} We extract the desired motion from the source skeleton sequence, then retarget the motion to the target person. Videos from the Internet vary drastically in body structure as shown in Fig. \ref{fig:motion_transfer}. For example, Spiderman has very long legs but the child has short ones. Our method, no matter how large the structural gap between the source and the target, is capable of generating a skeleton sequence precisely with the same body structure as the target person while preserving the motion from the source person. \\
\noindent\textbf{Novel-view synthesis.} We can explicitly manipulate the view of decoded skeleton in the 3D space, rotating it before projecting it down to 2D. We show an example in Fig. \ref{fig:novel_view}. This enables us to see the motion-transferred video at any desired view-angle. \\
\noindent\textbf{Latent space interpolation.} The learned latent representation is meaningful when interpolated, as shown in Fig.~\ref{fig:interpolation}. Both the motion and the the body structure change smoothly between the videos, demonstrating the effectiveness of our model in capturing a reasonable coverage of the manifold.


\begin{table}[h]
\caption{\textbf{Quantitative Results.} \textit{MSE} and \textit{MAE} are joint position errors measured on Mixamo, reported in the original scale of the data. \textit{FID} measure the quality of rendered images. \textit{User}s evaluate the consistency between source videos and generated videos. We report the percentage of users who prefer our model and our in-the-wild trained model, respectively.}
\vspace{-0.45cm}
\begin{center}
\resizebox{1.0\linewidth}{!}{
\small
\begin{tabular}{c|cc|c|cc}
\Xhline{1.2pt}
Method & MSE & MAE & FID & User & User (wild) \\ 
\Xhline{1.2pt}
LN  & 0.0886 & 0.1616 & 48.37 & 81.7\% & 82.9\% \\
NKN~\cite{NKN} & 0.0198 & 0.0781 & 67.32 & 84.5\% & 86.3\% \\
EDN~\cite{EDN} & 0.1186 & 0.2022 & 40.56 & 75.2\% & 77.1\% \\ 
LCM~\cite{LCM} & 0.0151 & 0.0749 & 37.15 & 68.5\% & 71.6\% \\
\Xhline{1.2pt}
\textbf{Ours} & \textbf{0.0131} & \textbf{0.0673} & \textbf{31.26} & - & - \\
\textbf{Ours (wild)} & \textbf{0.0121} & \textbf{0.0627} & \textbf{31.29} & - & - \\
\Xhline{1.2pt}

\end{tabular}
}
\end{center}
\label{tab:comparison_baseline}
\vspace{-0.45cm}
\end{table}

\subsection{Comparisons to State-of-the-Art Methods}

We compare the motion retargeting results of our method with the following methods (including one intuitive method and three state-of-the-art methods) both quantitatively and qualitatively.
1) \textbf{Limb Normalization} is an intuitive method that calculates a scaling factor for each limb and applies local normalization.
2) \textbf{Neural Kinematic Networks (NKN)} \cite{NKN} uses detected 3D keypoints for unsupervised motion retargeting.
3) \textbf{Everybody Dance Now (EDN)} \cite{EDN} applies a global linear transformation on all the keypoints.
4) \textbf{Learning Character-Agnostic Motion (LCM)} \cite{LCM} performs disentanglement at the 2D space in a fully-supervised manner.

For the fairness of the comparison, we train and test all the models on a unified Mixamo dataset, but note that our model is trained with less information, using neither 3D information~\cite{NKN} nor the pairing between motion and skeletons~\cite{LCM}. In addition, we train a separate model with in-the-wild data only. All the methods are evaluated with the aforementioned evaluation metrics.



Our method outperforms all the compared methods in terms of both numerical joint position error and quality of generated image. EDN and LN are naive rule-based methods, the former does not estimate the body structure and the latter is bound to fail when the actor is not facing the camera directly. Despite that NKN is able to transfer motion with little error on the synthesized dataset, it suffers on in-the-wild data due to the unreliability of 3D pose estimation. LCM is trained with a finite set of characters, therefore its capacity of generalization is limited. In contrast, our method uses limb-scaling to augment the training data, exploring all possible body structures in a continuous space.


It is noteworthy that our method enables training on arbitrary \textit{web data} that previous methods are not able to. The fact that the model trained on in-the-wild data (\ie, Solo-Dancer Dataset) achieved the lowest error (in Table~\ref{tab:comparison_baseline}) demonstrates the benefits of training on in-the-wild data. For complex motion such as the one shown in Fig.~\ref{fig:comparison_wild}, the model learned from wild data performs better, as wild data features a larger diversity of motion. These results show the superiority of our method in learning from unlimited real-world data, while supervised methods rely on strictly paired data that are hard to expand.



In summary, we attribute the superior performance of our method to the following reasons: 1) Our disentanglement is directly performed in 2D space, which circumvents the imprecise process of 3D-keypoints detection from in-the-wild videos. 2) Our explicit invariance-driven loss terms maximize the utilization of information contained in the training data, evidenced by the largely increased data efficiency compared to implicit unsupervised approaches~\cite{NKN}. 3) Our limb scaling mechanism improves the model's ability to handle extreme body structures. 4) In-the-wild videos provide an unlimited source of motion, compared to limited movements in synthetic datasets like Mixamo~\cite{mixamo}.


\begin{table}[h]
\caption{\textbf{Ablation Study Results.}}
\vspace{-0.45cm}
\begin{center}
\resizebox{0.95\linewidth}{!}{
\small
\begin{tabular}{c|ccc|c}
\Xhline{1.2pt}
Method & w/o crs & w/o trip & w/o adv  & Ours (full) \\
\Xhline{1.2pt}
MSE & 0.0392 & 0.0154 & 0.0136 & 0.0131 \\
MAE & 0.1259 & 0.0708 & 0.0682 & 0.0673 \\
\Xhline{1.2pt}
\end{tabular}
}
\end{center}
\label{tab:ablation}
\vspace{-0.45cm}
\end{table}

\subsection{Ablation study}

We train some ablated models to study the impact of the individual loss terms. The results are shown in Table~\ref{tab:ablation}.
We design three ablated models. The \emph{w/o crs} model is created by removing the cross reconstruction loss. The \emph{w/o trip} model is created by removing the triplet loss. The \emph{w/o adv} model is created by removing the adversarial loss. Removing the cross reconstruction loss has the most detrimental effect to the 2D retargeting performance of our model, evidenced by the doubling of MSE. Removal of the triplet loss increased the MSE by about $16\%$. Although removing the adversarial loss does not significantly affect the 2D retargeting performance of our model, the rotated sequences look less natural without it.


\section{Conclusion}
In this work, we propose a novel video motion retargeting approach, in which motion can be successfully transferred in scenarios where large variations of body-structure exist between the source and target person. The proposed motion retargeting network runs on 2D skeleton input only, makes it a lightweight and plug-and-play module, which is complementary to existing skeleton extraction and skeleton-to-video rendering methods. Leveraging three inherent invariance properties in temporal sequences, the proposed network can be trained with unlabeled web data end-to-end. Our experiments demonstrate the promising results of our methods and the effectiveness of the invariance-driven constraints. 

\noindent
\textbf{Acknowledgement.} This work is supported by the SenseTime-NTU Collaboration Project, Singapore MOE AcRF Tier 1 (2018-T1-002-056), NTU SUG, and NTU NAP. We would like to thank Tinghui Zhou, Rundi Wu and Kwan-Yee Lin for insightful discussion and their exceptional support.

{\small
\bibliographystyle{ieee_fullname}
\bibliography{egbib}
}

\clearpage
\section*{Appendix}

The content of our supplementary material is organized as follows.

\begin{enumerate}
    \item Details about the implementation of the three stages of our method. 
    \item Datasets and evaluation metrics we use in our experiments.
    \item Qualitative results of ablation study.
\end{enumerate}
\subsection*{S1. Implementation details}
\label{sec:1}

\subsubsection*{S1.1 Skeleton Extraction}
We use a pretrained DensePose model \cite{DenseposeGithub} for skeleton extraction, missing keypoints are complemented by nearest-neighbor interpolation. The extracted skeleton sequences are smoothed using a gaussian kernel with a temporal standard deviation $\sigma=2$. We use $N=15$ joints for a skeleton, detailed skeleton format will be given in our Github repository.

\subsubsection*{S1.2 Motion Retargeting Network}

The sizes of the latent representations are $C_m=128$, $C_s=256$ and $C_v=8$. Our encoders down-sample the input sequences to an eighth of its original length, therefore $M=\frac{T}{8}$. For limb-scaling, we use global and local scaling factors randomly sampled from $[0.5, 2]$ (uniformly distributed). For view perturbations we use $K=3$. Our motion retargeting network is trained $200,000$ steps with batch size $64$ and learning rate $\alpha=0.0002$ using Adam \cite{kingma2014adam} optimization algorithm. The weights of the loss terms are given as follows: $\lambda_\text{rec}=10, \lambda_\text{crs}=4, \lambda_\text{adv}=2, \lambda_\text{trip}=10, \lambda_\text{inv}=2$. These parameters are determined through quantitative and qualitative experiments on a validation set.

\subsubsection*{S1.3 Skeleton-to-Video Rendering}
For skeleton-to-video rendering, we recorded target videos of $5$ subjects as training data (none of the recorded subjects is an author of this work). We use the synthesis pipeline proposed in \cite{EDN}. Each generator is trained on the target video for $40$ epochs and the output size is $512 \times 512$. 

\begin{figure}[t]
\begin{center}
\includegraphics[width=1.0\linewidth]{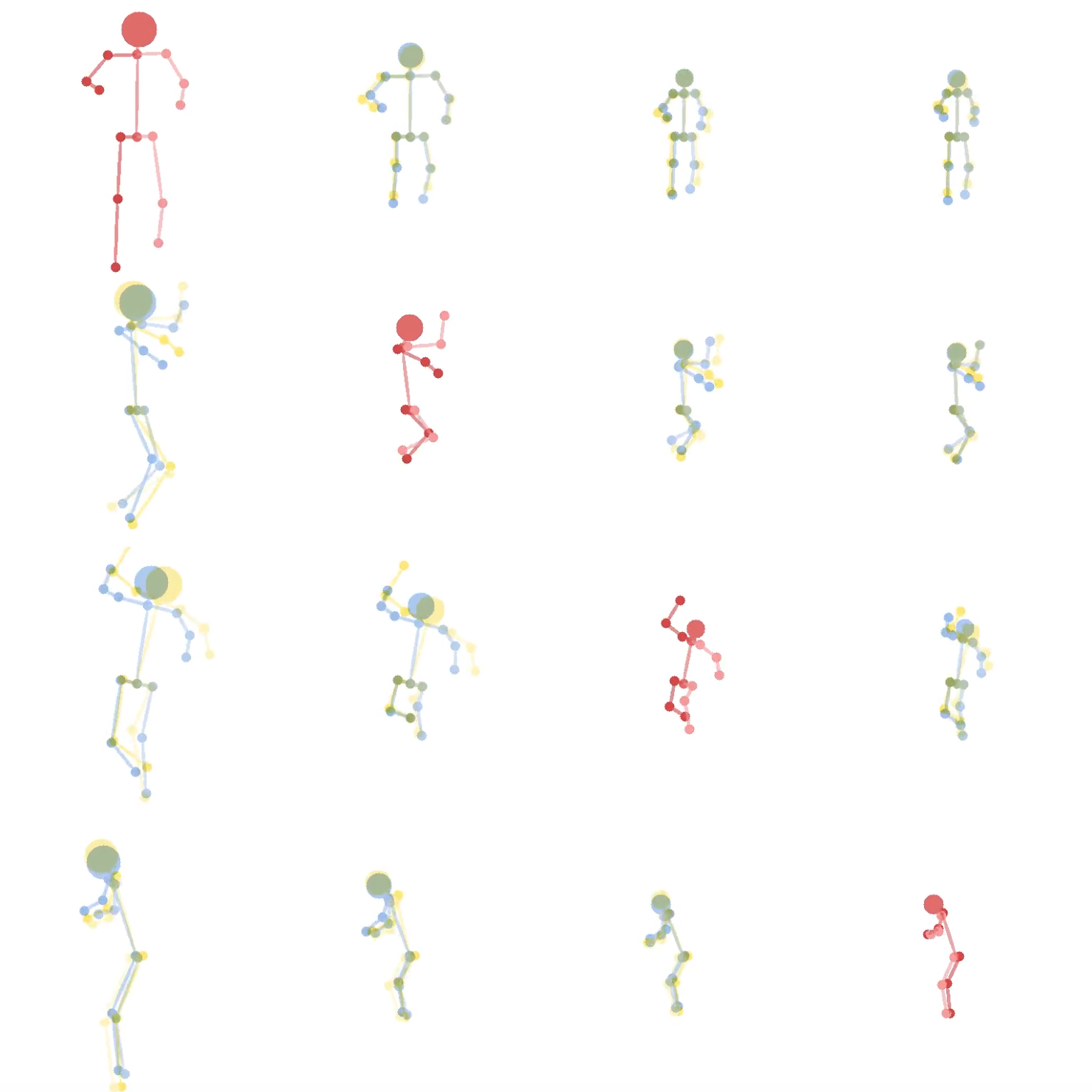}
\end{center}
\caption{\textbf{Visualization of retargeting error computation with our model.} In this figure, we plotted input joint sequences (red) on the diagonal. Off the diagonal are the retargeted sequences (blue) from our model as well as the groud truth (yellow), where their overlapping areas become green. In this figure, sequences on the same row are expected to perform the same motion, while sequences on the same column are expected to share the same body structure.}
\label{fig:comparison_gt}
\end{figure}

\subsection*{S2. Experimental Details}
\label{sec:2}

\subsubsection*{S2.1 Dataset}
\paragraph{In-the-wild dataset.}
For training on unlabeled web data, we collected a motion dataset named Solo-Dancer. We downloaded from YouTube $8$ categories of $337$ dancing videos, each one of the videos features only a single dancer. The total length of the videos add up to $11.5$ hours. We then used an off-the-shelf 2D keypoints detector \cite{alp2018densepose} to extract keypoints frame-by-frame to be used as our training data.
\paragraph{Sythesized dataset.}
We also perform the proposed unsupervised training pipeline on the synthetic Mixamo dataset \cite{mixamo} in order to quantitatively measure the transfer results with ground truth and baseline methods. The training set comprises of $32$ characters, each character has $800$ sequences and a total of $1.2$ hours for each character. 

\subsubsection*{S2.2 Evaluation Metrics}

\paragraph{MSE and MAE.}

For an inferred sequence $\hat{\mathbf{x}}$ and a groundtruth sequence $\mathbf{x}$
\[\text{MSE} = \frac{1}{2NT}\sum_{i, t} (\mathbf{x}_{i,t} - \hat{\mathbf{x}}_{i, t})^2 \]
where $i$ is the subscript of body joints and $t$ is the subscript of time. Similarly,
\[\text{MAE} = \frac{1}{2NT}\sum_{i, t} |\mathbf{x}_{i,t} - \hat{\mathbf{x}}_{i, t}| \]
These two metrics are measured in the original scale of Mixamo dataset. The errors are computed after hip-alignment, as visualized in Figure \ref{fig:comparison_gt}.

\paragraph{FID.}

We calculate the Frechet Inception Distance (FID)~\cite{heusel2017gans} to evaluate the quality of generated frames. FID measures the perceptual distance between the generated frames and the real target frames, and smaller number represents higher visual consistency.

\paragraph{User study.}

For the quality of retargeted videos, we ask $100$ volunteers to perform subjective pairwise A/B tests. For each method ($4$ baseline and $2$ ours), we test $90$ retargeted videos with the combination of $30$ source and $3$ target individuals. All the videos are 10 seconds in length. Participants choose which video has better motion consistency (between source videos and retargeted videos) in a pair of retargeted videos from two different methods. Source videos are also given to testers for reference. For each baseline method, $90$ retargeted videos are compared $100$ times by different participants against \textit{our model}. Our model has two variants with different training sets (\ie, Mixamo and SoloDancer), the results are shown in Table 1 in main paper as ``User'' and ``User (wild)'' respectively.

\begin{figure}[t]
\begin{center}
\includegraphics[width=1.0\linewidth]{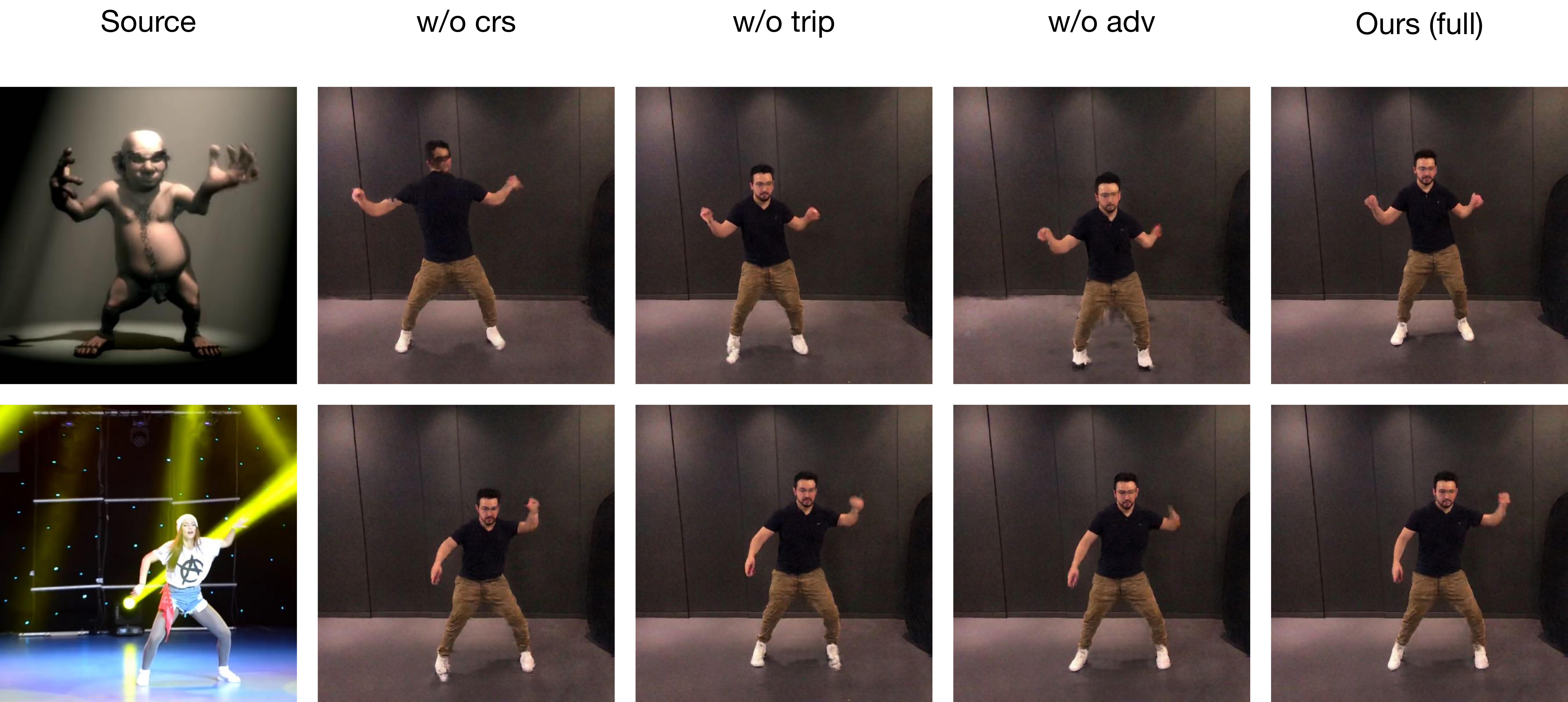}
\end{center}
\caption{\textbf{Qualitative results of ablation study.} The first column gives the motion sources, and the other columns show corresponding results.}
\label{fig:supp_ablation}
\end{figure}

 \subsection*{S3. Qualitative Ablation Study}
 \label{sec:3}
 
 Besides testing standard MSE, we render the retargeted video for further comparison. As can be empirically observed in Fig. \ref{fig:supp_ablation}, the full model produces the results of the best quality. The cross reconstruction loss plays an essential role for disentanglement. The results without triplet loss show slightly degraded quality on the frame level. However, it is important to note that the triplet loss is used to smooth the structure and view code temporally, therefore stabilizing the generated video. The adversarial loss improves the plausibility of generated joint sequences, making them look more natural and realistic. Recall that the adversarial loss is added on randomly rotated output joint sequences to make the rotated output sequences indistinguishable from real data. 
 




\end{document}